\documentclass[letterpaper,10pt,journal]{IEEEtran}

\usepackage{graphics}
\usepackage{epsfig}
\usepackage{times}
\usepackage{amsmath}
\usepackage{amssymb}
\usepackage{cite}
\usepackage{orcidlink}
\usepackage{mathtools}
\usepackage{bm}
\usepackage{xcolor}
\usepackage{algorithm}
\usepackage{algpseudocode}
\usepackage{svg}
\usepackage{placeins}
\usepackage{hyperref}
\usepackage{subcaption}
\usepackage{booktabs}
\usepackage{multirow}
\usepackage{array}
\title{A Physics-Informed Collision Learning Framework for Collaborative Robot Motion Generation}

\author{Chen Cai\orcidlink{0009-0002-3474-6671} and Steven Liu\orcidlink{0000-0002-3407-2745}\thanks{Chen Cai and Steven Liu are with
	the Department of Electrical and Computer Engineering, University of Kaiserslautern Landau,
	Kaiserslautern, 67663, Germany (e-mail: \{chen.cai, steven.liu\}@rptu.de).}\thanks{ This work has been submitted to the IEEE for possible publication. Copyright may be transferred without notice, after which this version may no longer be accessible.}}
\hypersetup{
hidelinks,
pdfauthor={Chen Cai and Steven Liu},
pdftitle={A Physics-Informed Collision Learning Framework for Collaborative Robot Motion Generation},
pdfsubject={Preprint submitted to IEEE Transactions on Automation Science and Engineering},
pdfkeywords={collaborative robots, collision avoidance, differentiable collision detection, motion planning, model predictive control, robot learning}
}

\begin{document}

\maketitle

\begin{abstract}
Close-proximity multi-arm manipulation requires collision models that are both geometrically accurate and differentiable enough for real-time optimization. Classical geometry checkers provide reliable distances but are difficult to use inside gradient-based model predictive control, while conservative proxy models can restrict tightly coupled motion. We present PI-UDF, a physics-informed unified differentiable framework for body-to-body collision distance prediction between articulated robots. PI-UDF combines analytical forward kinematics with learnable link-geometry embeddings and a shared residual network to predict pairwise inter-arm distances directly from robot configurations. To improve safety-critical fidelity, we combine
quota-driven boundary mining with an asymmetric
boundary-crossing penalty that emphasizes false-safe sign
errors near the collision boundary. The learned distance field is integrated into nonlinear MPC as a differentiable inter-arm clearance term. We validate the framework on a real dual-Franka platform through high-speed close-proximity 14-DoF dual-arm swapping, sustained single-arm dynamic evasion, and dynamic-evasion planning configurations with frozen, predicted, and target-switching treatments of the moving arm. Hardware experiments and offline Drake/FCL replay show that PI-UDF provides a differentiable inter-arm clearance estimate suitable for closed-loop collision-aware collaborative robot motion generation.
\end{abstract}


\section{INTRODUCTION}
Close-proximity multi-arm manipulation requires multiple robots to move through tight shared workspaces while maintaining continuous inter-arm clearance. Classical
mesh-based checkers such as FCL~\cite{pan2012fcl} provide
reliable geometry queries for offline validation, but their
discrete and non-smooth computations are difficult to embed in gradient-based model predictive control. Real-time
controllers often replace full geometry with
simplified convex proxies; for example, distributed MPC can
approximate robot links by line segments and ellipsoids~\cite{gafur2022dynamic}. These proxies reduce computation,
but their inflated clearances can restrict tightly coupled
motion and cause deadlock-like behavior in dense
workspaces~\cite{zhu2025efficient}.

Signed distance fields provide a continuous clearance signal
and gradients that are useful for optimization-based collision
avoidance. Neural SDF variants extend this idea to robot
configuration spaces; for example, Neural-JSDF uses MLPs
for joint-space distance fields~\cite{koptev2022neural}, while
Bernstein-polynomial representations improve storage
efficiency~\cite{liu2023collision}. However, most such
formulations remain centered on point-to-robot queries rather
than direct body-to-body clearance between two moving
articulated arms.

For multi-arm planning, treating one moving arm as a set of
query points for the other arm requires dense online point
sampling or proxy-shape approximations. Both choices scale
poorly with link complexity and weaken the geometry signal
available to a closed-loop optimizer. What is needed is a
compact, differentiable body-to-body distance model that
reasons directly over robot configurations.

\begin{figure*}[t]
  \centering
    \begin{subfigure}{0.38\textwidth}     \centering
    \includegraphics[width=\linewidth]{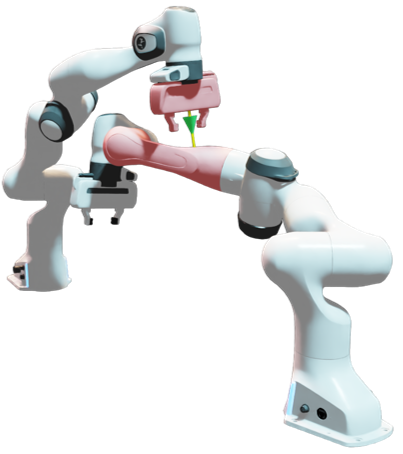}
    \caption{Body-to-Body SDF}
  \end{subfigure}
  \hfill
  \begin{subfigure}{0.31\textwidth}     \centering
    \includegraphics[width=\linewidth]{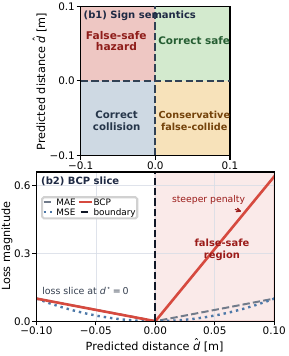}
    \caption{Asymmetric BCP Loss}
  \end{subfigure}
  \hfill
  \begin{subfigure}{0.27\textwidth}     \centering
    \includegraphics[width=\linewidth]{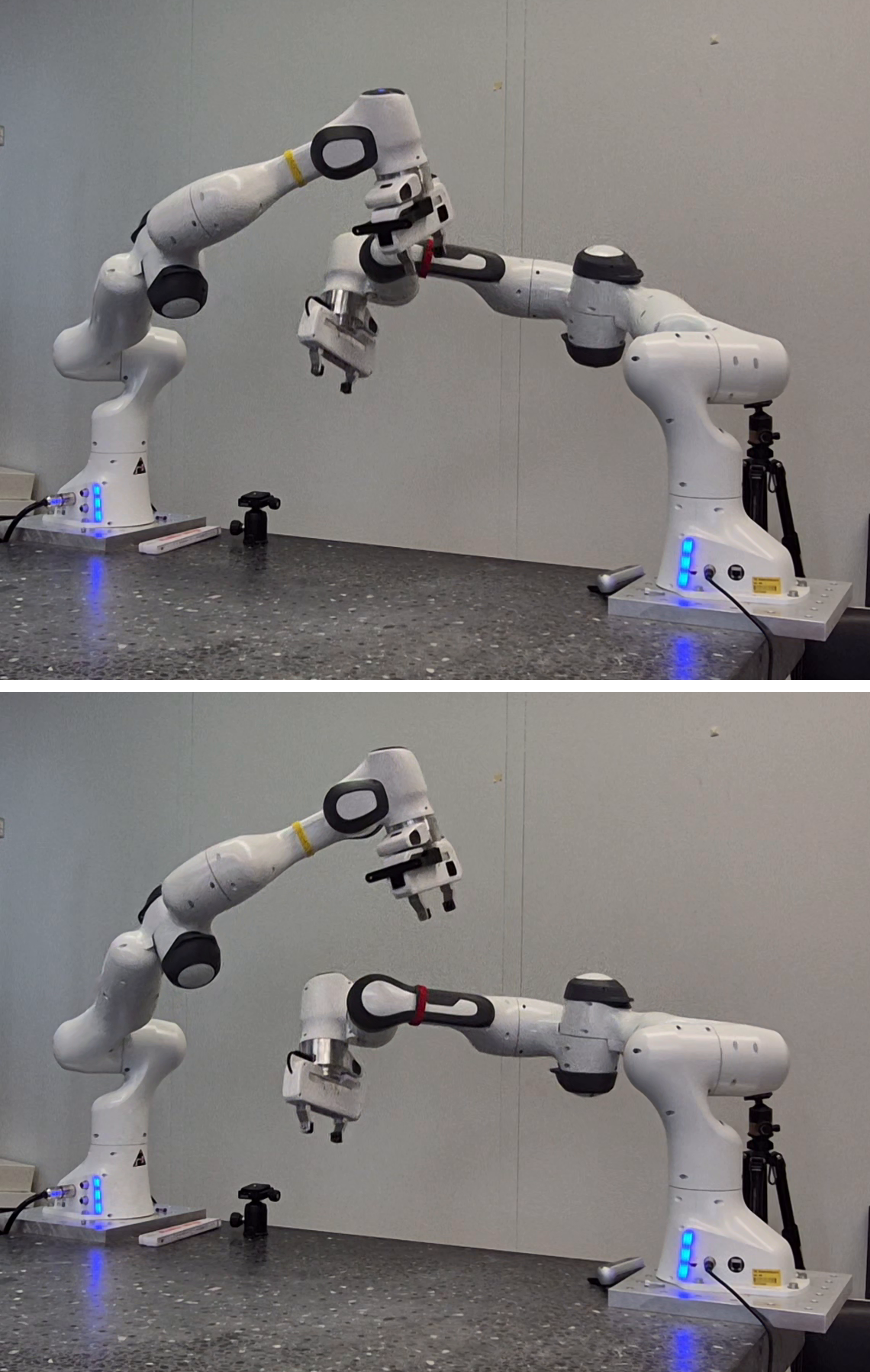}
    \caption{Closed-loop motion generation}
  \end{subfigure}
  
    \caption{\textbf{Overview of the PI-UDF framework.} (a) Physics-informed Body-to-Body SDF provides differentiable inter-arm clearance and usable gradients for optimization. (b) The Boundary-Crossing Penalty (BCP) emphasizes
    false-safe boundary violations while keeping conservative
    safe-state errors less costly. (c) The learned distance field is used as an inter-arm clearance term in NMPC for close-proximity motion generation. Representative frames during close-proximity dual-arm crossing and subsequent separation are shown.}
    \label{fig:teaser}
    \end{figure*}

PairwiseNet~\cite{kim2023pairwisenet} is an important step
toward direct link-pair distance learning. It decomposes the
global collision problem into pairwise subproblems and
conditions a shared regressor on pairwise relative transforms
and point-cloud-derived shape descriptors, thereby avoiding
dense online point sampling during planning. Our framework builds
on this established pairwise relative-pose factorization and
focuses its contribution on the co-design of geometry
representation, boundary-focused supervision, and
differentiable integration into downstream motion generation.

Boundary-focused data acquisition has also been studied for
high-DoF collision-distance learning. Kim and
Park~\cite{kim2024active} iteratively update a fixed-size
configuration dataset with samples near the current model's
global collision boundary and use a redundant link-
\(SE(3)^n\) representation to improve distance learning.
Their approach is model-in-the-loop: the learned collision
boundary guides subsequent data generation. In contrast,
our retained-set construction is driven by ground-truth
coverage deficits over link-pair classes and distance zones,
rather than by the boundary of an intermediate model.

Against this background, we propose the Physics-Informed
Unified Differentiable Framework (PI-UDF), a compact
framework for body-to-body collision distance learning
between articulated robots. PI-UDF provides a differentiable
collision-distance representation that can be queried by
downstream motion-generation modules; in this work, we
instantiate this interface within nonlinear MPC. The model
conditions a shared residual regressor on FK-derived relative
poses and task-optimized link embeddings learned end-to-end
from signed-distance supervision. PI-UDF further combines
pair-level boundary mining with an asymmetric
Boundary-Crossing Penalty that emphasizes false-safe
prediction errors.

The contributions of this work are threefold:

\begin{itemize}

\item \textbf{Task-optimized geometry representation with
analytical kinematics.}
We condition a shared residual regressor on FK-derived
relative poses and task-optimized link embeddings learned
end-to-end from signed-distance supervision. This replaces
a separately pretrained point-cloud representation with
lightweight embedding lookups, yielding a compact
\(0.11\,\mathrm{M}\)-parameter model and an approximately
\(5\times\) wall-clock training speedup over PairwiseNet at
the same 50-epoch budget.

\item \textbf{Pair-level boundary supervision with asymmetric
safety-aware learning.}
We introduce ground-truth class-zone quota mining at the
link-pair level together with an asymmetric Boundary-Crossing
Penalty that emphasizes false-safe sign errors. This improves
coverage of penetration and near-contact interactions while
reducing hazardous boundary-crossing errors.

\item \textbf{Differentiable motion-generation integration and hardware validation.}
We integrate PI-UDF into a JIT-compiled nonlinear MPC stack
as a differentiable inter-arm clearance model and validate the
complete learning--control pipeline on real dual-Franka
hardware through close-proximity swapping, dynamic evasion,
and reactive/predictive operating modes, with offline
Drake/FCL replay for geometry-level validation.

\end{itemize}

\section{Physics Informed Unified Collision Learning Framework}
\label{sec:PIUDF Framework}
\subsection{Problem Formulation: Pairwise Decomposition with
Analytical Kinematics}

Consider a collaborative system comprising two manipulators
with joint configurations
$q_1 \in \mathbb{R}^{n_1}$ and $q_2 \in \mathbb{R}^{n_2}$.
We denote the full dual-arm configuration by
$q=[q_1^\top,q_2^\top]^\top \in \mathbb{R}^{n_1+n_2}$.
Let $\mathcal{V}_r(q_r) \subset \mathbb{R}^3$ be the
workspace volume occupied by robot $r \in \{1,2\}$ at
configuration $q_r$. The inter-arm signed clearance is defined as
\begin{equation}
                        \mathcal D^\star(q)
    =
    \operatorname{sdist}
    \bigl(\mathcal V_1(q_1),\mathcal V_2(q_2)\bigr).
    \label{eq:global_body_distance}
\end{equation}
where \(\operatorname{sdist}(\cdot,\cdot)\) denotes the signed
set-to-set clearance: positive for separated bodies, zero at
contact, and negative under penetration. Directly learning $\mathcal{D}^\star(q)$ as a
monolithic map from the full joint space is data inefficient:
the network would have to infer both the rigid-body kinematic
transformations and the local geometry-dependent distance
field from sparse high-dimensional samples.

PI-UDF avoids this entanglement by using the known
kinematic chain as an analytical prior. Let
$\mathcal{G}_{r,i}$ denote the fixed local geometry of link $i$
on robot $r$, and let
${}^{0}T_{r,i}(q_r)=\mathrm{FK}_{r,i}(q_r) \in SE(3)$ be the
world-frame pose of that link computed by forward kinematics.
For an inter-arm link pair $(i,j)$, we define the relative pose
\begin{equation}
    T_{ij}(q)
    =
    \left({}^{0}T_{1,i}(q_1)\right)^{-1}
    {}^{0}T_{2,j}(q_2)
    \in SE(3).
    \label{eq:relative_pose}
\end{equation}
The global body-to-body clearance can then be written as the
minimum over pairwise signed distances:
\begin{equation}
\mathcal D^\star(q)
=
\min_{(i,j)\in\mathcal P}
d_{ij}^\star
\bigl(T_{ij}(q),G_{1,i},G_{2,j}\bigr).
    \label{eq:pairwise_decomposition}
\end{equation}
where $\mathcal{P}$ is the set of inter-arm link pairs considered
for collision reasoning, and $d_{ij}^{\star}$ is the ground-truth
signed distance between the two link geometries. Equation~(3) decomposes the global inter-arm clearance into
link-pair subproblems. Within each subproblem, analytical
forward kinematics computes the configuration-dependent
relative pose \(T_{ij}(\mathbf q)\), while the learned
regressor models its interaction with the fixed link
geometries. This separates the computational roles of
kinematics and geometry without treating them as
independent.

The learning problem is therefore reduced from fitting a
global joint-space distance field to approximating a family of
shape-conditioned pairwise distance functions. PI-UDF
represents the fixed link geometries by learnable descriptors
\(\mathbf{e}_{1,i}\) and \(\mathbf{e}_{2,j}\), obtained from
robot-specific embedding tables, and uses a single shared
regressor to predict
\begin{equation}
    \hat d_{ij}(q)
    =
    f_\theta
    \!\left(
        T_{ij}(q),
        \mathbf{e}_{1,i},
        \mathbf{e}_{2,j}
    \right)
    \approx
    d_{ij}^{\star}(q).
    \label{eq:piudf_pairwise_regressor}
\end{equation}
Analytical kinematics provides the relative transform, leaving
the neural network to learn the remaining
geometry-conditioned distance mapping.

For evaluation and reporting, the raw predicted inter-arm
clearance is
\begin{equation}
    \hat{\mathcal D}(q)
    =
    \min_{(i,j)\in\mathcal P}
    \hat d_{ij}(q).
    \label{eq:raw_predicted_clearance}
\end{equation}
Because the exact minimum is non-smooth when the active
link pair changes, NMPC instead uses the smooth surrogate
\(\widetilde{\mathcal D}_{\alpha}(q)\) defined in Sec.~IV.

\subsection{Unified Dual-Stream Architecture}
\begin{figure*}[t]
	\centering
	\includegraphics[width=0.98\linewidth]{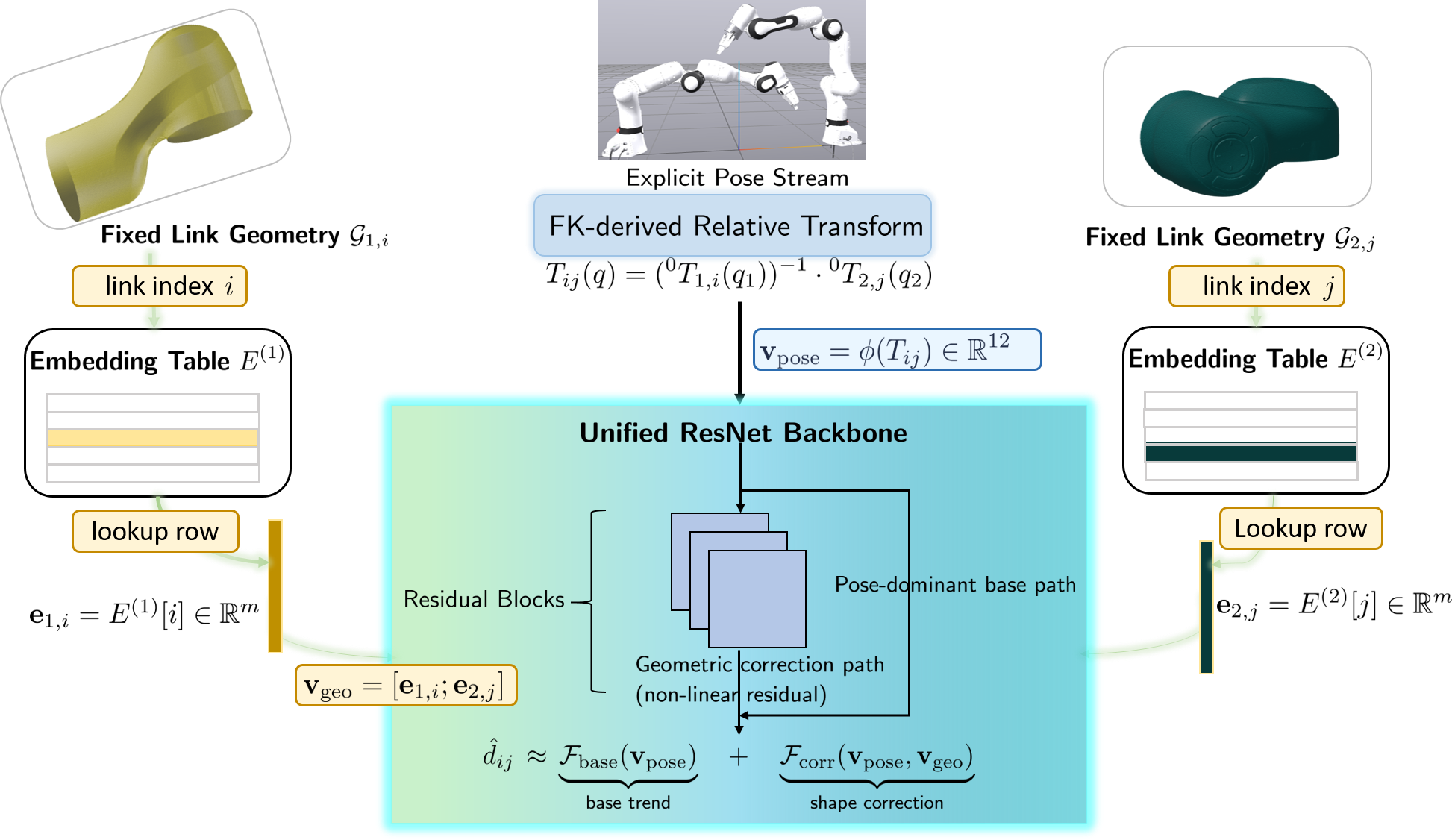}
		\caption{\textbf{Overview of the PI-UDF architecture.} Given an inter-arm link pair \((i,j)\), analytical forward
kinematics computes the world-frame link poses
\({}^{0}T_{1,i}(q_1)\) and \({}^{0}T_{2,j}(q_2)\), from which
the relative transform \(T_{ij}(q)\) is constructed and
vectorized as \(\mathbf{v}_{\mathrm{pose}}\). The fixed link geometries are represented by robot-specific learnable embedding-table lookups \(\mathbf{e}_{1,i}=E^{(1)}[i]\) and \(\mathbf{e}_{2,j}=E^{(2)}[j]\), forming the implicit geometry descriptor \(\mathbf{v}_{\mathrm{geo}}\). A shared residual regressor fuses both streams to predict the pairwise signed distance \(\hat d_{ij}\).}
	\label{fig:dualstream_pipeline}
	\end{figure*}

To instantiate the pairwise regressor in
Eq.~\eqref{eq:piudf_pairwise_regressor}, PI-UDF uses a
unified dual-stream architecture, as shown in
Fig.~\ref{fig:dualstream_pipeline}. The network input is composed
of an explicit pose descriptor derived from the analytical relative
transform \(T_{ij}(q)\), and an implicit geometry descriptor obtained
from learnable link-embedding tables. The two descriptors are fused
by a shared residual regressor to predict the pairwise signed distance
\(\hat d_{ij}\).

\textbf{Explicit pose stream.}
For each inter-arm link pair \((i,j)\), the relative pose
\(T_{ij}(q)\in SE(3)\) is already provided by Eq.~\eqref{eq:relative_pose}.
We convert it into a vector representation
\[
    \mathbf{v}_{\mathrm{pose}}
    =
    \phi(T_{ij})
    \in \mathbb{R}^{12},
\]
where \(\phi(\cdot)\) flattens the rotation matrix and translation
vector of \(T_{ij}\). This descriptor exposes the relative spatial
configuration of the two links directly to the network, instead of
requiring the network to infer forward kinematics from joint angles.
Because the predictor uses only \(T_{ij}\), it is invariant
to a common rigid transformation applied to both links.

\textbf{Implicit geometry stream.}
The link geometries \(\mathcal{G}_{1,i}\) and \(\mathcal{G}_{2,j}\)
are fixed for a given robot model and therefore need not be encoded
from point clouds during inference. We represent them using
robot-specific learnable embedding tables
\[
    E^{(r)} \in \mathbb{R}^{L_r \times m}, \qquad r\in\{1,2\},
\]
where \(L_r\) is the number of modeled links of robot \(r\), and
\(m=32\) in our implementation. For a query pair \((i,j)\), the
geometry descriptors are obtained by table lookup:
\[
    \mathbf{e}_{1,i}=E^{(1)}[i], \qquad
    \mathbf{e}_{2,j}=E^{(2)}[j],
\]
and concatenated as
\[
    \mathbf{v}_{\mathrm{geo}}
    =
    [\mathbf{e}_{1,i};\mathbf{e}_{2,j}]
    \in \mathbb{R}^{2m}.
\]
These embedding rows are initialized randomly and optimized jointly
with the distance regressor using the collision-distance supervision.
Thus, the learned descriptors are not intended to reconstruct visual
shape, but to encode geometry features that are useful for signed
distance prediction. During inference, this geometry stream reduces
to constant-time indexing. Identical robots may share the same
embedding table, while heterogeneous robot pairs can be represented
by robot-specific tables without changing the shared regressor
architecture.

The fused feature vector
\[
    \mathbf{z}_{ij}
    =
    [\mathbf{v}_{\mathrm{pose}}^\top,
     \mathbf{v}_{\mathrm{geo}}^\top]^\top
\]
is passed to a shared residual backbone. We interpret the residual predictor as
\begin{equation}
    \hat d_{ij}
    =
    f_\theta(\mathbf{z}_{ij})
    \approx
    \underbrace{\mathcal{F}_{\mathrm{base}}
    (\mathbf{v}_{\mathrm{pose}})}_{\text{pose-dominant trend}}
    +
    \underbrace{\mathcal{F}_{\mathrm{corr}}
    (\mathbf{v}_{\mathrm{pose}},\mathbf{v}_{\mathrm{geo}})}
    _{\text{geometry-conditioned correction}} .
    \label{eq:residual_distance_model}
\end{equation}
Equation~(6) provides a conceptual additive interpretation
of the learned mapping rather than an explicitly implemented
two-branch architecture, reflecting the structure of the distance query:
the relative translation and orientation already provide a strong
low-order cue for inter-link clearance, while the remaining correction
depends on the detailed link shapes. The shared backbone therefore
does not relearn rigid-body kinematics, but concentrates its capacity
on geometry-conditioned deviations from the pose-dominant distance
trend. Since the same \(f_\theta\) is reused for all link pairs, the
model avoids both a heavy pretrained point-cloud encoding and a separate expert
network for each pair. Moreover, when the experimental platform uses two identical Franka robots, the implementation shares a single embedding table between the two arms.

\subsection{Safety-Aware Training Pipeline}
Representative neural collision-learning methods rely on
broad configuration-space sampling
\cite{kim2023pairwisenet,zhu2025efficient}, while PairwiseNet
additionally adopts a standard symmetric regression objective
\cite{kim2023pairwisenet}. In a 14-DoF dual-arm system,
however, contact and near-contact configurations occupy only
a small, structured subset of the joint space. Uniform
\(q\)-sampling therefore produces predominantly far-field
data and severely underrepresents the boundary interactions
most relevant to closed-loop safety. We refer to this
imbalance as the \textbf{Sparsity Trap}.

Model-in-the-loop active learning has previously addressed
this sparsity by iteratively updating a fixed-size dataset with
configurations sampled near the learned global collision
boundary~\cite{kim2024active}. Our objective differs in two
respects. First, PI-UDF is supervised on pairwise distances
rather than a single global configuration-level label. Second,
our mining rule is based on ground-truth coverage deficits
over link-pair classes and distance zones, without repeatedly
estimating and resampling around an intermediate model
boundary.

The imbalance is amplified at the link-pair level because
each configuration induces many \(d_{ij}^{\star}\) labels:
a globally safe configuration may contain informative
near-field pairs, whereas a boundary configuration may
contain only one or two critical pairs. PI-UDF therefore
retains physically valid whole configurations while performing
mining and supervision at the link-pair level. Its
\textbf{Safety-Aware Training Pipeline} combines
class-zone quota-driven active mining with an asymmetric
boundary-crossing loss, reflecting that neither all samples
nor all errors are equally important.

\subsubsection*{1) Quota-Driven Link-Pair Active Mining}

We distinguish the configuration-level clearance
\(\mathcal D^\star(q)\) from the pairwise signed distance
\(d_{ij}^{\star}(q)\). The global collision state of a dual-arm
configuration is determined by
\[
    \mathcal D^\star(q)
    =
    \min_{(i,j)\in\mathcal P}d_{ij}^\star(q).
\]
whereas PI-UDF is supervised on individual inter-arm link-pair
queries. Each retained configuration can be expanded into samples of
the form
\[
    s = \big(q,i,j,T_{ij}(q),d_{ij}^{\star}(q)\big),
\]
where \(T_{ij}(q)\) is the analytical relative pose defined in
Eq.~\eqref{eq:relative_pose}. The distance zone used for mining and
batch construction is assigned by the pairwise label
\(d_{ij}^{\star}(q)\), not by the configuration-level ground-truth clearance
\(\mathcal D^\star(q)\).

For fair comparison with joint-space baselines, we use the same
budget of 100k retained dual-arm configurations. The difference is
how these configurations are selected. Uniform sampling passively
accepts most far-field configurations, whereas our active generator
retains a candidate configuration only when its induced link-pair
distances improve the coverage of underrepresented boundary regions.
Specifically, candidate configurations are proposed in joint space and
evaluated over the canonical inter-arm link-pair set
\(\mathcal{P}\). Each pairwise query is assigned to one of four
distance zones:
\[
\begin{aligned}
    W_1 &: d_{ij}^{\star}\le 0
    &&\text{(penetration)},\\
    W_2 &: 0<d_{ij}^{\star}\le 5\,\mathrm{cm}
    &&\text{(near-field)},\\
    W_3 &: 5\,\mathrm{cm}<d_{ij}^{\star}\le 20\,\mathrm{cm}
    &&\text{(mid-field)},\\
    W_4 &: d_{ij}^{\star}>20\,\mathrm{cm}
    &&\text{(far-field)}.
\end{aligned}
\]

For each link-pair class and each zone, we maintain a prescribed
coverage quota. A candidate configuration is retained if at least one of its link-pair queries contributes to an underrepresented
class-zone bucket; otherwise it is rejected. This converts the fixed
configuration budget into a boundary-enriched set of pairwise
supervision signals. In practice, higher priority is assigned to
kinematically relevant and historically under-covered pairs, such as
distal forearm, wrist, and end-effector interactions, because these
pairs dominate close-proximity collaborative manipulation.

Retaining a configuration preserves labels for all canonical
inter-arm link pairs. Thus, a configuration accepted because
one pair lies near contact also supplies mid- and far-field
labels for the remaining pairs, preserving global calibration.
Mini-batches are sampled from the resulting class-zone
buckets, while Fig.~\ref{fig:zone_dist} summarizes the
retained set at the configuration level through
\(\mathcal{D}^{\star}(q)\).
\subsubsection*{2) Asymmetric Boundary-Crossing Penalty}

Active mining determines which regions of the distance field are
sampled frequently; the loss determines which errors are treated as
safety-critical. Standard MSE penalizes far-field and boundary errors
symmetrically. For MPC, however, the critical failure mode is a
\emph{false-safe} prediction, where the model predicts positive
clearance although the ground-truth pair is in contact or penetration.

To suppress this failure mode, we use an asymmetric
Boundary-Crossing Penalty (BCP). Let \(d_s^{\star}\) and
\(\hat d_s\) denote the ground-truth and predicted signed distances
of a pairwise training sample \(s\), and let
\(z(s)\in\{W_1,W_2,W_3,W_4\}\) be its distance-zone label. We define
\(I_s=\mathbb{I}[\hat d_s>0 \land d_s^{\star}\le0]\) as the
false-safe boundary-crossing indicator. The loss is
\begin{equation}
\label{eqa: penalty_func}
\mathcal{L}_{\mathrm{BCP}}
=
\frac{1}{N}
\sum_{s=1}^{N}
\lambda_{z(s)}
\left|d_s^{\star}-\hat d_s\right|
\left(1+\eta I_s\right).
\end{equation}

Here, \(\eta\) controls the strength of the asymmetric safety wall:
only false-safe boundary crossings are amplified, while conservative
safe-state errors remain penalized by the base weighted \(L_1\) term.
The zone weight \(\lambda_{z(s)}\) further prioritizes boundary
regions, with larger weights for penetration and near-field samples,
e.g.,
\(\lambda_{W_1},\lambda_{W_2}>\lambda_{W_3}>\lambda_{W_4}\).
Thus, PI-UDF concentrates supervision and loss pressure on
safety-critical link-pair interactions while retaining mid- and
far-field coverage for global calibration.

\section{Network Evaluation and Ablation Studies}
\label{sec:network_model}
\subsection{Experimental Setup and Training Settings}
\label{sec:exp_setup}

To rigorously evaluate the proposed framework, we structure our experiments to assess both system-level safety and the underlying architectural choices.

\subsubsection{Target System}
As illustrated in Fig.~\ref{fig:teaser}(a), we evaluate PI-UDF on a
fixed-cell dual-arm system consisting of two 7-DoF Franka Emika Panda
arms with a fixed relative base placement (inter-base offset
\(1.16\,\mathrm{m}\), opposing orientations \([0,\pi]\) rad). This setup
matches the intended deployment scenario of many industrial
collaborative workcells, where the robot bases are calibrated once and
remain fixed during operation.

This choice should be distinguished from the original goal of
PairwiseNet~\cite{kim2023pairwisenet}, which targets generalized
collision detection across varying multi-arm layouts. Although the analytical relative transform already includes
the robot-base placement, training across multiple layouts
would be required to cover the resulting relative-pose
distribution. Here we focus on the calibrated fixed-cell
deployment. This fixed-cell setting allows us to evaluate whether
a compact pairwise differentiable distance model can provide accurate near-boundary distances and gradients for downstream motion planning.

\subsubsection{Training Datasets}
\label{sec:dataset_mining}
    \begin{figure}[htbp]
        \centering
        \includegraphics[width=0.98\linewidth]{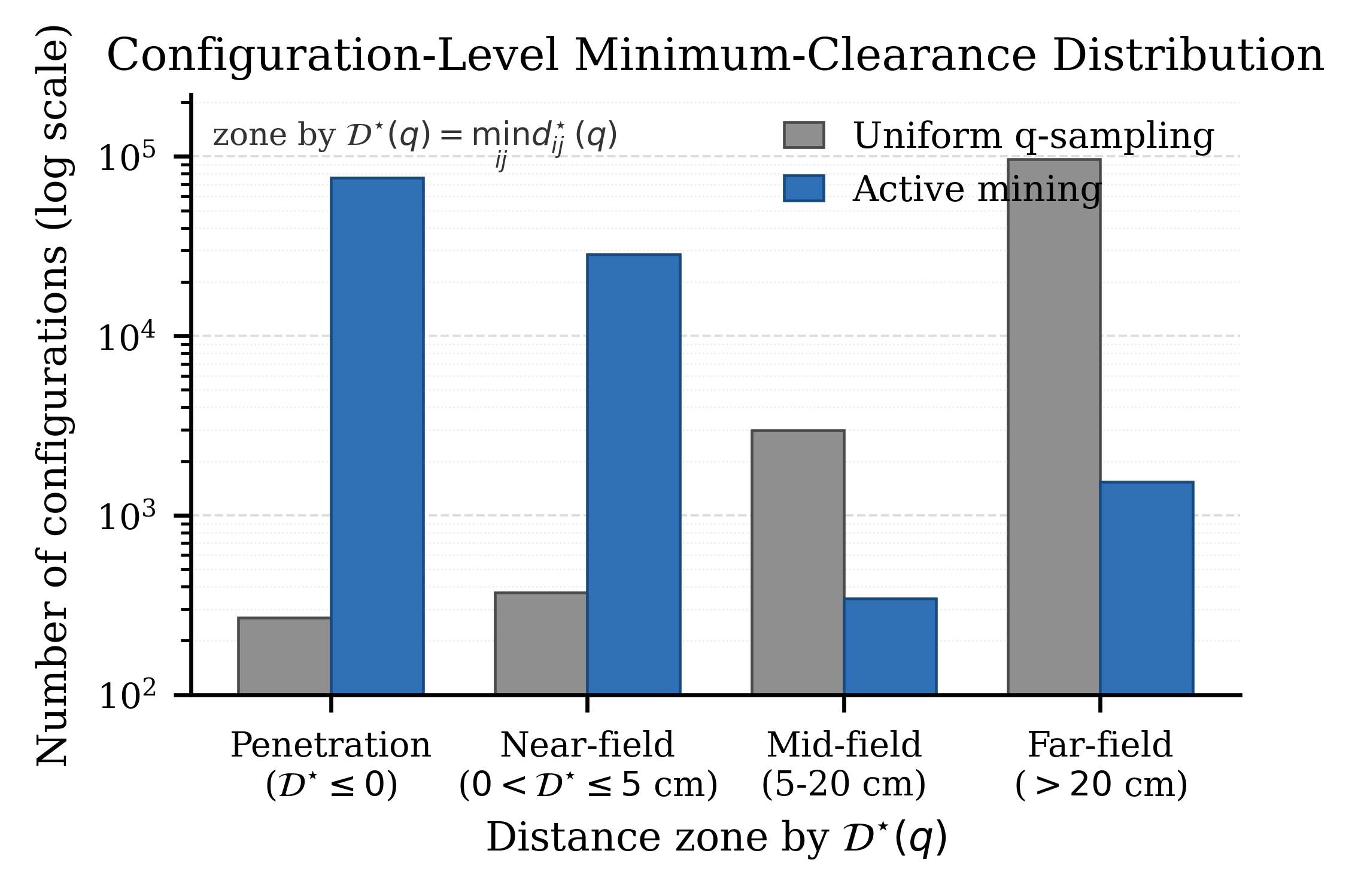}
        \caption{\textbf{Configuration-level distance-zone distribution.}
        Each retained dual-arm configuration is assigned to a zone by its
        minimum inter-arm signed clearance
        \(\mathcal{D}^{\star}(q)=
        \min_{(i,j)\in\mathcal{P}} d_{ij}^{\star}(q)\).
        Under the same \(100\mathrm{k}\) retained-configuration budget,
        uniform \(q\)-sampling is dominated by far-field configurations,
        whereas our quota-driven active mining shifts the retained set toward
        penetration and near-field configurations. The link-pair-level quota
        mechanism used to construct this retained set is described in Sec.~II-C.}
        \label{fig:zone_dist}
    \end{figure}

We compare uniform and active datasets under the same
budget of \(100\,\mathrm{k}\) retained dual-arm
configurations. The uniform set follows the joint-space
sampling used in our PairwiseNet reproduction, whereas the
active set is constructed using the class-zone quota rule in
Sec.~II-C. Figure~\ref{fig:zone_dist} summarizes the induced
configuration-level coverage using
\[
\mathcal{D}^{\star}(q)
=
\min_{(i,j)\in\mathcal P} d_{ij}^{\star}(q).
\]
Only \(0.64\%\) of uniformly sampled configurations fall in
the penetration or near-field zones, compared with over
\(98\%\) in the actively retained set. Both retained sets are
then expanded into canonical link-pair queries for training;
thus, the active-data PairwiseNet variants isolate the effect
of boundary coverage from the effects of the PI-UDF
representation and BCP loss. Although the active set is
boundary-focused at the configuration level, each retained
configuration still provides mid- and far-field labels for
non-critical link pairs.

\subsection{Implementation and Baselines}
\label{sec:implementation}

\subsubsection{Experimental Infrastructure}
\label{sec:hardware specs}
All data generation, training, and evaluation procedures were executed on a standard workstation equipped with an Intel Core i7-12700 CPU (2.10\,GHz, 12 cores), 64\,GB DDR4 RAM, and an NVIDIA GeForce RTX 5080 GPU. The proposed PI-UDF and all baseline models were implemented in PyTorch. For training, we generally utilize the Adam optimizer with a learning rate of $1\times 10^{-4}$ and a batch size of $512$. Unless stated otherwise, BCP-trained variants use
\(\eta=5.0\),
\((\lambda_{W_1},\lambda_{W_2},\lambda_{W_3},\lambda_{W_4})
=(1.5,1.0,0.5,0.1)\), and balanced \(4{:}4{:}4{:}4\)
zone sampling.

\subsubsection{Baseline Variants}
We use PairwiseNet~\cite{kim2023pairwisenet} as the primary
pairwise-learning baseline. Because pretrained
geometry-encoder weights were unavailable, we trained the
point-cloud encoders for the same Franka link geometries
using the authors' public pipeline and froze them during
subsequent distance-regressor training. All PairwiseNet
results below therefore denote a budget-matched fixed-cell reproduction on the same
fixed-base dual-Franka setup. We evaluate the following
variants:

\begin{itemize}
    \item \textbf{PairwiseNet (Uniform, 100 ep):}
    The PairwiseNet regressor is trained on the uniformly
    sampled joint-space dataset, following the original
    sampling paradigm, for \(100\) epochs.

    \item \textbf{PairwiseNet (Active, 50/75 ep):}
    The same PairwiseNet encoder--regressor architecture is
    trained on our boundary-enriched active dataset for
    \(50\) and \(75\) epochs. These variants isolate the
    effect of data coverage while preserving the PairwiseNet
    representation and symmetric regression objective.

    \item \textbf{PI-UDF (Active+BCP, 50 ep):}
    The complete PI-UDF model, comprising FK-derived
    relative-pose inputs, jointly learned link embeddings,
    the SiLU ResNet backbone, and BCP training, is optimized
    on the active dataset for \(50\) epochs.
\end{itemize}

Beyond these system-level benchmarks, a rigorous dissection of PI-UDF's internal architectural choices -- specifically the impact of the ResNet backbone and continuous activation functions -- is detailed in the subsequent ablation studies (Sec.~\ref{sec:ablation}).

\subsubsection{Held-Out Static and Dynamic Test Sets}
\label{sec:test_sets}
To rigorously evaluate the models' worst-case safety performance and practical reliability, we construct two distinct test domains. 

First, we generate a Static Safety-Critical Test Set via rejection sampling from an independently sampled dual-arm configurations not used
during training. We filter randomly drawn dual-arm configurations to enforce a strict quota: 30\% true collisions ($d \le 0$), 50\% near-collisions ($0 < d \le 5\,\text{cm}$), and 20\% safe states in terms of minimum inter-arm collision distance. The final static set comprises 2,000 accepted configurations (yielding 110,000 valid inter-arm link-pair evaluations), with ground-truth signed distances computed via Drake (FCL/GJK) \cite{drake}.

Second, to assess temporal consistency, we define a Dynamic Trajectory Test Set using a high-speed dual-arm ''swapping'' motion. By continuously evaluating the raw predicted clearance \(\hat{\mathcal D}(q_t)\) against the Drake/FCL reference \(\mathcal D^\star(q_t)\), we rigorously benchmark their point-wise accuracy and robustness against prediction chattering.
\subsection{Comparative Evaluation on Held-Out Test Sets}
\label{sec:system_perf}

Building upon the held-out test sets defined previously, we benchmark the model performance across several metrics:
\begin{itemize}
    \item \textbf{RMSE (cm):} The root mean-squared error (RMSE) between the estimated and actual collision distances.
    \item \textbf{SC-MAE (cm):} Safety-Critical Mean Absolute Error targets precision exclusively within the narrow near-field and collision zones ($d < 5$\,cm).
    \item \textbf{FNR (\%):} The False Negative Rate represents the frequency of predicting a physically colliding state ($d^* \le 0$) as safe ($d > 0$). 
    \item \textbf{AUC:} Area Under the ROC Curve. Evaluates the overall classification capability of distinguishing safe versus colliding states across all continuous thresholds.
\end{itemize}

\subsubsection*{Evaluation on the Static Safety-Critical Test Set}
\begin{figure}[t]
  \centering
  \includegraphics[width=\linewidth]{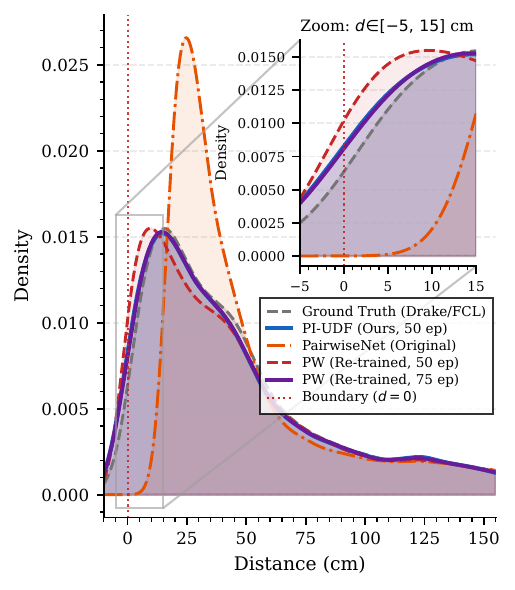}
  \caption{\textbf{Static Prediction Density.} Probability density (KDE) of distance predictions on the static safety-critical set. PI-UDF captures the ground-truth penetration boundary ($d \le 0$), whereas PairwiseNet (Original) shifts toward false-safe values, reflecting poor near-field/penetration field pre-training.}
  \label{fig:pred_dist}
\end{figure}

Table~\ref{tab:master_perf} and Fig.~\ref{fig:pred_dist} detail the models' performance on the static test set, offering insights into the efficacy of our specific methodological contributions:

\noindent\textbf{Dataset Quality and Sparsity Trap:} 
Under the reproduced uniform-data setting, PairwiseNet
yields an FNR of \(100\%\), with its predicted density mass shifted erroneously into the false-safe positive region (Fig.~\ref{fig:pred_dist}). Retraining the same regressor
on the active dataset reduces the FNR to \(16.47\%\),
isolating the effect of boundary-oriented data coverage.

\begin{table*}[t]
\caption{System-Level Performance and Computational Efficiency.
           Evaluated on the Static Safety-Critical Test Set ($N{=}110{,}000$). Best results are \textbf{bold}; second-best are \underline{underlined}. PI-UDF time includes end-to-end training of its embeddings and regressor. PairwiseNet times report distance-regressor training with a frozen geometry encoder and exclude the one-time encoder-pretraining cost. All reported times are measured on the same workstation.
          }
  \label{tab:master_perf}
  \centering
  \renewcommand{\arraystretch}{1.2}
  \begin{tabular}{lrrrrrr}
    \toprule
    \textbf{Model}
      & \textbf{Params (M)} $\downarrow$
      & \textbf{Train Time} $\downarrow$
      & \textbf{RMSE} $\downarrow$
      & \textbf{SC-MAE} $\downarrow$
      & \textbf{FNR} $\downarrow$
      & \textbf{AUC} $\uparrow$ \\
      & \scriptsize{(Trainable / Frozen)} 
      & \scriptsize{(min)} 
      & \scriptsize{(cm)} 
      & \scriptsize{(cm)} 
      & \scriptsize{(\%)} 
      & \\
    \midrule
    PairwiseNet (Uniform, 100\,ep)          & \underline{0.04 / 0.81} & ${\approx}270.0$ & 8.70 & 18.33          & 100.00          & 0.9221 \\
    PairwiseNet (Active, 50\,ep) & \underline{0.04 / 0.81} & \underline{138} & 4.80 & 2.38           & 16.47           & 0.9888 \\
    PairwiseNet (Active, 75\,ep) & \underline{0.04 / 0.81} & 202.5            & \textbf{3.08} & \textbf{2.27}  & \underline{7.48}& \textbf{0.9950} \\
    \midrule
    \textbf{PI-UDF (Ours; Active+BCP, 50\,ep)}   & \textbf{0.11 / 0.00}    & \textbf{27.8}    & \underline{3.32} & \underline{2.33} & \textbf{6.83}   & \underline{0.9948} \\
    \bottomrule
  \end{tabular}
\end{table*}

\noindent\textbf{Loss Formulation and Safety:} 
When comparing models at an equal training budget of 50 epochs, PairwiseNet (Re-trained) and PI-UDF achieve highly comparable SC-MAE ($2.38$\,cm vs.\ $2.33$\,cm). This suggests that our enriched dataset levels the playing field for near-field precision. Yet, PI-UDF achieves a substantially lower FNR ($6.83\%$ vs.\ $16.47\%$). This safety advantage is consistent with the proposed
Boundary-Crossing Penalty (BCP), which amplifies false-safe
boundary errors, i.e., predicting positive clearance when the
ground-truth pair is in contact or penetration. Compared with the
symmetric loss used by the re-trained PairwiseNet regressor, BCP
therefore reduces hazardous sign errors while preserving comparable
near-field accuracy.

\noindent\textbf{Global Regression vs. Training Efficiency:} 
Because BCP and the zone weights place less emphasis on far-field
errors than on boundary errors, one might expect PI-UDF's global RMSE to degrade. PI-UDF (50\,ep) maintains a RMSE of $3.32$\,cm, closely trailing the larger PairwiseNet trained for $1.5\times$ longer (75\,ep, RMSE $3.08$\,cm). As observed in the far right of Fig.~\ref{fig:pred_dist}, PI-UDF still closely approximates the ground-truth density in the safe zone($d > 20$\,cm). This confirms that our framework attains competitive global regression and lowest FNR among the compared models, without requiring a massive pre-trained geometric encoder or exhaustive training cycles.

\subsubsection*{Continuous Dynamic Trajectory Tracking}
\begin{figure*}[t]   \centering
  \includegraphics[width=0.93\linewidth]{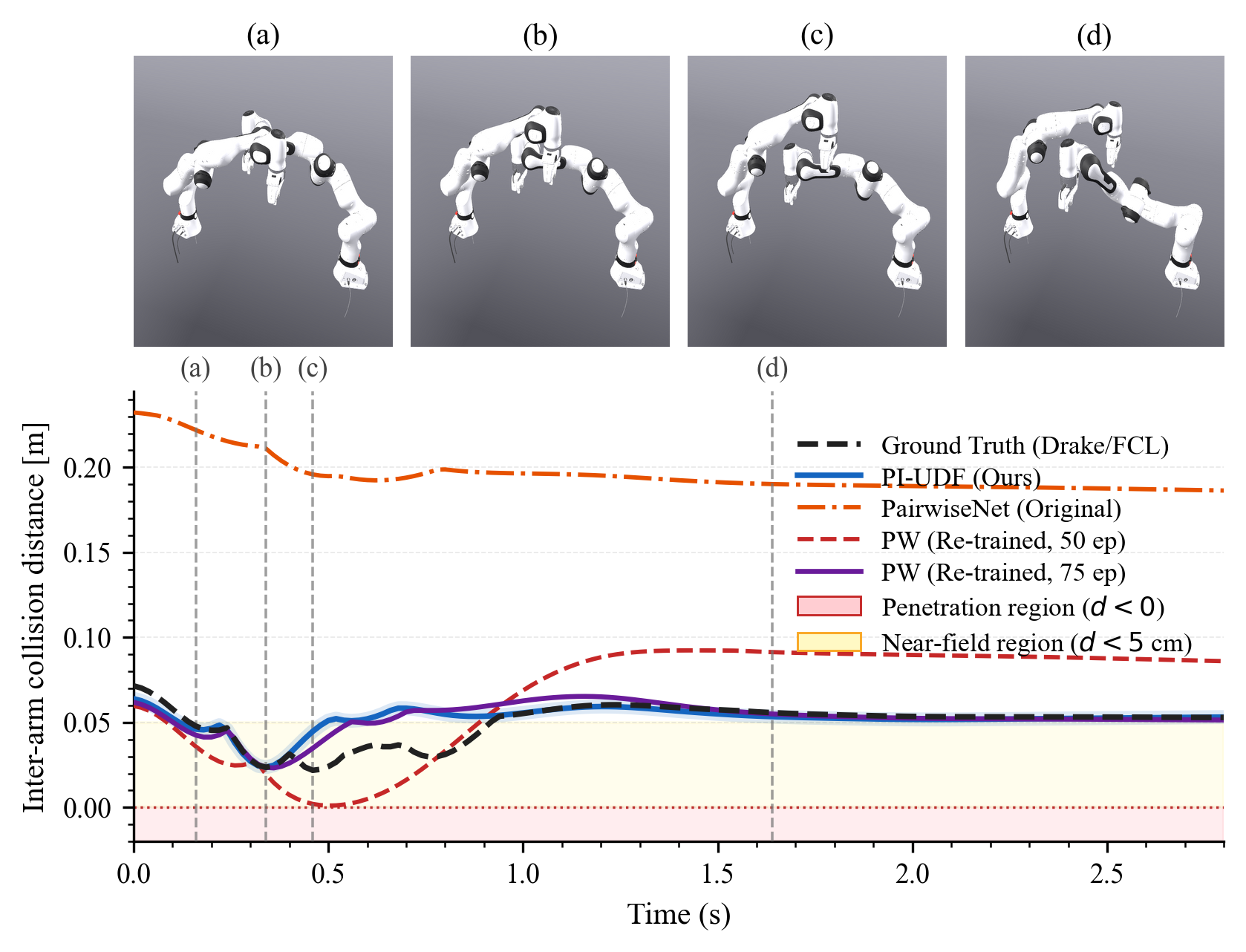}
  \caption{\textbf{Continuous Dynamic Trajectory Tracking.} Raw predicted
minimum clearances \(\hat{\mathcal D}(q_t)\) are compared
with the Drake/FCL reference \(\mathcal D^\star(q_t)\)
during the high-speed swapping trajectory.}
  \label{fig:dynamic_traj}
\end{figure*}

Static metrics alone cannot guarantee the temporal consistency required for closed-loop control. We evaluate continuous predictions during a high-speed swapping maneuver in Fig.~\ref{fig:dynamic_traj}. PI-UDF maintains a smooth, continuous prediction profile that tightly overlaps with the ground-truth collision trend across all spatial zones, exhibiting minimal error in near field. The PairwiseNet (Re-trained, 75\,ep) baseline also demonstrates similar trend tracking. In contrast, while the modestly trained PairwiseNet (Re-trained, 50\,ep) manages to capture the general motion trend, it exhibits noticeably larger predictive errors. 

\subsubsection*{End-to-End Architectural Efficiency}
Table~I further highlights the computational benefit of the unified
PI-UDF design. PairwiseNet relies on a frozen point-cloud encoder
with $0.81\,\mathrm{M}$ parameters in addition to its trainable regressor, whereas PI-UDF trains a compact $0.11\,\mathrm{M}$ parameter model end-to-end. Because the geometry representation is learned jointly with the distance regression objective and reused across all link pairs, PI-UDF avoids both online geometric encoding and link-specific expert networks. At the same 50-epoch training budget, this compact model achieves a lower FNR than the re-trained PairwiseNet baseline while reducing training time from 138 min to 27.8 min.

\subsection{Ablation Study}
\label{sec:ablation}

We ablate activation (ReLU/SiLU), backbone and capacity
(MLP/ResNet), and training strategy (MSE/BCP and
Base/FT/HNM) on the static safety-critical test set. All
comparisons follow the evaluation protocol in
Sec.~\ref{sec:test_sets}, and the late-stage variants use the
same total epoch budget. Because PI-UDF is differentiated
inside NMPC, we additionally report a trajectory-level joint-gradient
variation diagnostic.

Along the fixed L4--L5 collision-crossing trajectory
\(\{\mathbf{q}^{(t)}\}_{t=1}^{T}\), let
\[
\mathbf{g}_t =
\nabla_{\mathbf{q}}\hat{d}(t).
\]
We define the gradient roughness as
\begin{equation}
  \mathcal{R}_g
  =
  \frac{1}{T-1}
  \sum_{t=1}^{T-1}
  \left\|
    \mathbf{g}_{t+1}-\mathbf{g}_{t}
  \right\|_2 .
  \label{eq:roughness}
\end{equation}
A lower \(\mathcal{R}_g\) indicates smaller adjacent-step
input-gradient variation under the same trajectory and
discretization. The complete comparison is reported in
Table~\ref{tab:ablation}.

\begin{table*}[t]
  \renewcommand{\arraystretch}{1.15}
  \centering
\caption{  Ablation study of architectural and training-loss choices. 
  Regression metrics (RMSE, SC-MAE, FNR) are evaluated on the Static Safety-Critical Test Set ($N{=}110{,}000$), while gradient roughness $\mathcal{R}_g$ is a model-level gradient diagnostic along a standardized trajectory. 
  $\downarrow$: lower is better. The best result in each column is \textbf{boldfaced}.
  $^\dagger$BCP = Boundary-Crossing Penalty; 
  $^\ddagger$FT = Fine-tuned on boundary-region samples; 
  HNM = Hard Negative Mining.
  Note: Latency is measured on the Intel i7-12700 CPU (Sec.~\ref{sec:hardware specs}) using a batch size of $B{=}64$ single-pair queries. Reported values are wall-clock averages over 200 runs after 20 warmup iterations.
}
  \label{tab:ablation}
  \setlength{\tabcolsep}{6.5pt}
  \begin{tabular}{l lcc r rrrr r}
    \toprule
    \multirow{2}{*}{\textbf{Variant}}
      & \multirow{2}{*}{\textbf{Backbone}}
      & \multirow{2}{*}{\textbf{Act.}}
      & \multirow{2}{*}{\textbf{Loss}}
      & \multirow{2}{*}{\textbf{Params}}
      & \textbf{RMSE}
      & \textbf{SC-MAE}
      & \textbf{FNR}
      & $\bm{\mathcal{R}_g}$
      & \textbf{Lat.} \\
    & & & & &
      \textbf{(cm)$\,\downarrow$} &
      \textbf{(cm)$\,\downarrow$} &
      \textbf{(\%)$\,\downarrow$} &
      \textbf{($\times$10$^{-3}$)$\,\downarrow$} &
      \textbf{(ms)} \\
    \midrule
    PI-UDF (MLP)
      & MLP (128$\to$64$\to$1)      & ReLU & MSE
      & 18{,}465
      & 4.027 & 2.354 & 7.89 & 12.74 & 0.048 \\
    PI-UDF (MSE)
      & ResNet ($w{=}128, b{=}3$)   & SiLU & MSE
      & 109{,}345
      & 3.442 & 2.283 & 7.29 & 2.31 & 0.209 \\
    PI-UDF (Small)
      & ResNet ($w{=}64, b{=}2$)    & SiLU & BCP$^\dagger$
      & 21{,}921
      & 4.220 & 2.588 & 7.71 & 2.85 & 0.108 \\
    PI-UDF (ReLU)
      & ResNet ($w{=}128, b{=}3$)   & ReLU & BCP$^\dagger$
      & 109{,}345
      & 3.635 & 2.278 & 7.52 & 9.83 & 0.196 \\
    \midrule
    \textbf{PI-UDF (Base)} $\star$
      & ResNet ($w{=}128, b{=}3$)   & SiLU & BCP$^\dagger$
      & 109{,}345
      & 3.316 & 2.328 & 6.83 & 2.14 & 0.177 \\
    PI-UDF (FT)$^\ddagger$
      & ResNet ($w{=}128, b{=}3$)   & SiLU & BCP$^\dagger$
      & 109{,}345
      & \textbf{3.148} & \textbf{2.275} & 6.69 & 1.98 & 0.189 \\
    PI-UDF (HNM)$^\ddagger$
      & ResNet ($w{=}128, b{=}3$)   & SiLU & BCP$^\dagger$
      & 109{,}345
      & 3.298 & 2.365 & \textbf{6.37} & \textbf{1.89}$^{\,\ast}$ & 0.194 \\
    \bottomrule
  \end{tabular}
  \\[2pt]
  \footnotesize
  $^{\,\ast}$The PI-UDF (HNM) $\mathcal{R}_g$ value is lower than PI-UDF (Base) due to the concentrated boundary-region training during Hard Negative Mining, which further smooths the constraint landscape near $d{=}0$.
\end{table*}

\subsubsection{Impact of Activation Function (SiLU vs.\ ReLU)}
\label{sec:ablation_activation}

For fixed link embeddings, a ReLU regressor is piecewise
affine in its network input. After composition with the
analytical FK map, activation-boundary crossings can still
induce abrupt changes in the joint-space gradient
\(\nabla_{\mathbf q}\hat d\). SiLU replaces this hard gating with a smooth
transition. Such derivative regularity is relevant to methods
based on repeated local linearization, including
SLSQP~\cite{kraft1988software} and
iSAM2~\cite{kaess2012isam2}, since abrupt gradient changes
can reduce the fidelity of local models and quasi-Newton
updates~\cite{nocedal2006numerical}. A smooth activation
alone, however, does not guarantee solver convergence.

\begin{figure}[tbp]
  \centering
  \includegraphics[width=\linewidth]{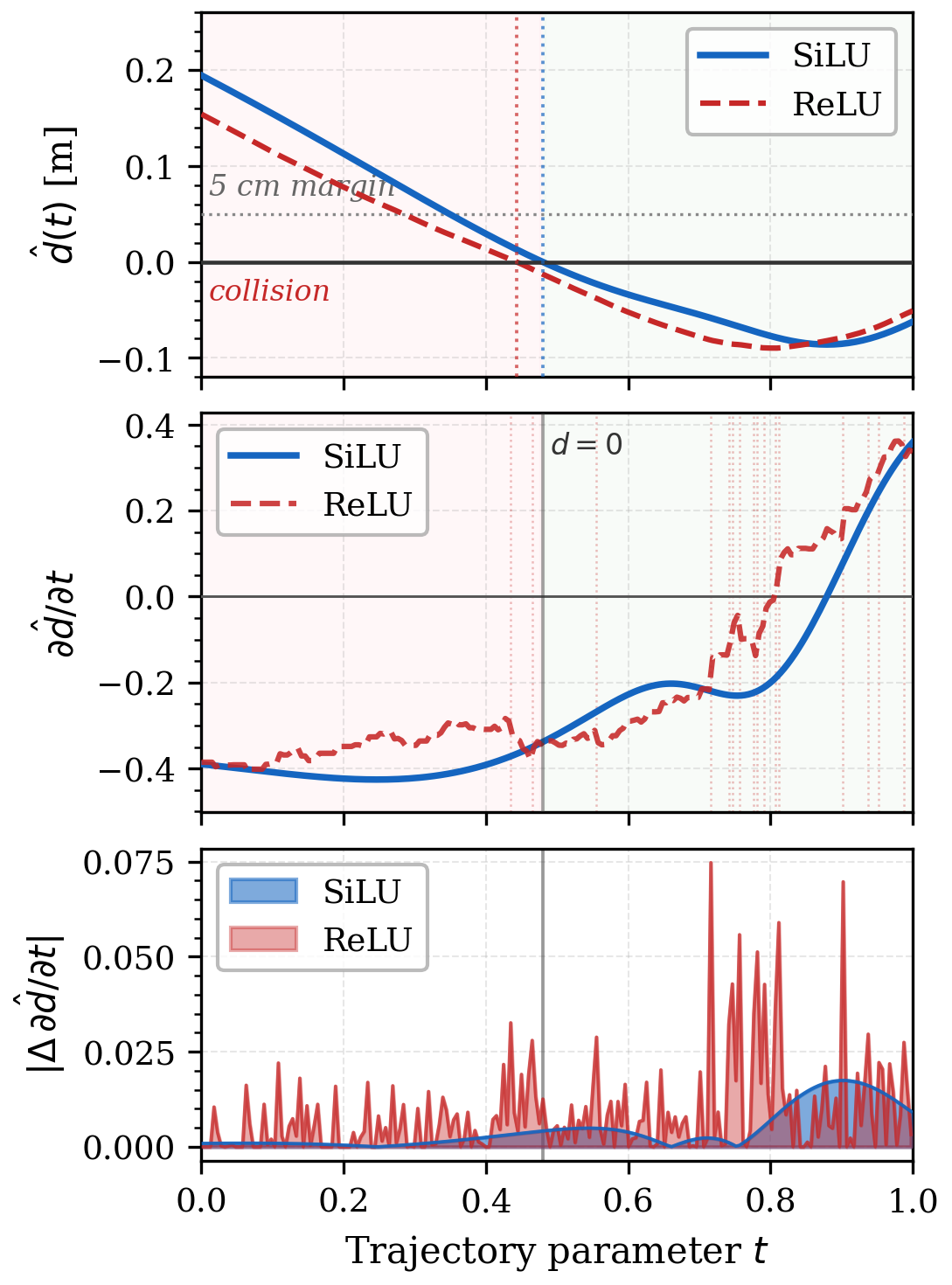}
  \caption{    Gradient smoothness along an SE(3) collision trajectory (link pair L4--L5). 
    \textbf{Top:} Predicted distance $\hat{d}$. 
    \textbf{Middle:} Directional derivative. ReLU exhibits piecewise-constant staircase artifacts, whereas SiLU remains $C^\infty$-smooth. 
    \textbf{Bottom:} Gradient roughness $|\Delta(\nabla \hat{d})|$. While ReLU's mean roughness is $1.95\times$ higher, its peak spikes are $\mathbf{4.28\times}$ more severe ($0.075$ vs.\ $0.017$) and occur deep within the collision region ($t{=}0.72$). These severe discontinuities can degrade local linearizations and quasi-Newton curvature estimates.
  }
  \label{fig:gradient_smoothness}
\end{figure}

We quantify this behavior using the model-level gradient
roughness \(\mathcal{R}_g\) in Eq.~\ref{eq:roughness} along
the fixed L4--L5 collision-crossing trajectory. Comparing
the matched \textbf{PI-UDF (ReLU)} and
\textbf{PI-UDF (Base)} variants, ReLU worsens RMSE from
\(3.316\) to \(3.635\,\mathrm{cm}\) and FNR from
\(6.83\%\) to \(7.52\%\), while SC-MAE remains comparable.
It also increases \(\mathcal{R}_g\) from
\(2.14\times10^{-3}\) to \(9.83\times10^{-3}\)
(\(4.6\times\)). Fig.~\ref{fig:gradient_smoothness} shows
the corresponding staircase-like directional derivative and
localized input-gradient spikes. We therefore use SiLU in
the deployed model.

ReLU's zero derivative for negative pre-activations may
also leave units persistently inactive~\cite{maas2013rectifier}.
SiLU avoids an identically zero negative branch, providing a
secondary motivation for signed-distance regression; our
activation choice is based primarily on the matched empirical
comparison above.

\begin{figure}[tbp]
  \centering
  \includegraphics[width=\linewidth]{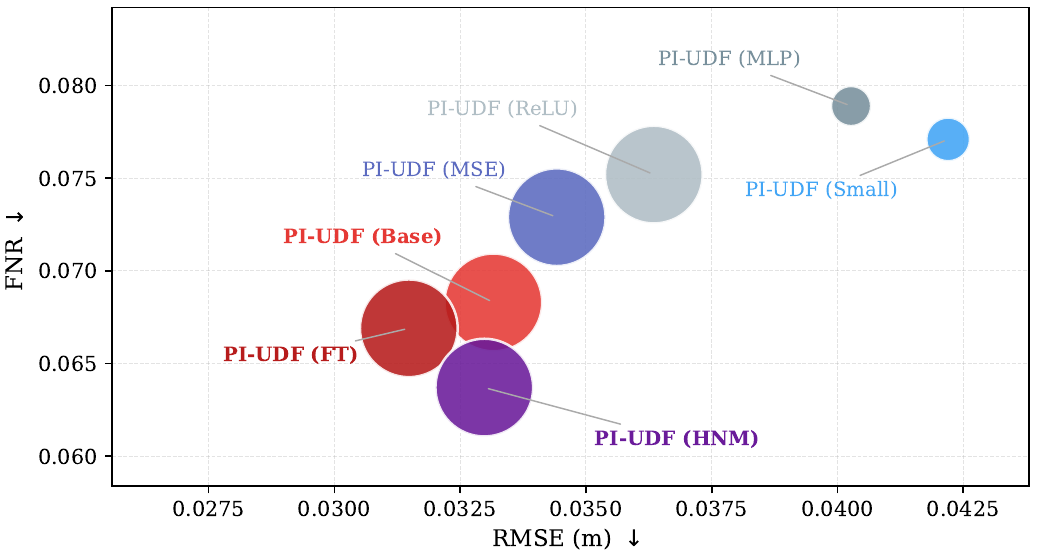}
  \caption{  RMSE--FNR trade-off across ablation variants
  (bottom-left is preferred); bubble area scales with parameter
  count. FT improves both metrics over Base, whereas HNM
  further lowers FNR with a small RMSE increase.
}
  \label{fig:ablation_bubble}
\end{figure}

\subsubsection{Backbone and Model Capacity}
\label{sec:ablation_arch}
The FK-derived relative-pose descriptor and learned link embeddings remain informative even with the shallow \textbf{PI-UDF (MLP)} baseline, which attains an RMSE of \(4.027\,\mathrm{cm}\) and an FNR of \(7.89\%\) (Table~\ref{tab:ablation}). This result shows that useful pairwise distance prediction can be obtained without a separately pretrained point-cloud geometry representation.

A matched capacity comparison is provided by \textbf{PI-UDF (Small)} and \textbf{PI-UDF (Base)}, which use the same SiLU activation and BCP loss. Increasing the residual backbone from \(21{,}921\) to \(109{,}345\) parameters reduces RMSE from \(4.220\) to \(3.316\,\mathrm{cm}\), SC-MAE from \(2.588\) to \(2.328\,\mathrm{cm}\), and FNR from \(7.71\%\) to \(6.83\%\). We therefore use the Base backbone for deployment. It remains compact at \(0.11\,\mathrm{M}\) parameters and evaluates a batch of \(64\) single-pair
queries in \(0.177\,\mathrm{ms}\) on an Intel i7-12700 CPU. Controller-level masking and smooth-min aggregation are described in Sec.~IV.

\subsubsection{Training Strategy: Boundary Fine-Tuning and HNM}
\label{sec:ablation_ft}

While \textbf{PI-UDF (Base)} establishes good accuracy-safety balance, its collision-awareness can be further customized via targeted late-stage training interventions. Rather than changing the architecture or increasing the computational budget, these stage-wise strategies modify the data distribution or loss objective after the initial epochs. Crucially, the total training epochs remains strictly constant across all variants, ensuring that performance gains stem entirely from the intervention mechanisms rather than extended training time.

As shown in Table~\ref{tab:ablation} and Fig.~\ref{fig:ablation_bubble}, standard boundary fine-tuning (\textbf{PI-UDF (FT)}) -- achieved by shifting the batch sampling distribution heavily toward contact boundaries ($|d| < 5$\,cm) -- improves both primary metrics, reducing RMSE to $3.148$\,cm and FNR to $6.69\%$. Alternatively, Hard Negative Mining (\textbf{PI-UDF (HNM)}) alters the loss landscape by explicitly up-weighting missed deep-collision configurations, i.e., false-safe predictions in penetration states. This heightened penalty further drives the FNR down to $6.37\%$, albeit introducing a slight conservative bias that marginally increases the global RMSE to $3.298$\,cm.

In summary, we present PI-UDF not merely as a static architecture, but as an adaptable framework. While the Base configuration serves as a strong default, the FT and HNM strategies demonstrate that practitioners can seamlessly steer the network in later epochs to satisfy specific deployment criteria. For instance, tasks demanding high spatial precision in tight clearances can leverage FT to maximize global accuracy, whereas highly safety-critical applications can employ HNM to enforce
stronger collision conservatism and suppress false-safe predictions.

\section{OPTIMIZATION-BASED Motion PLANNING INTEGRATION}
\label{sec:nmpc}
\subsection{PI-UDF-NMPC Formulation}
\label{sec:piudf_nmpc}
PI-UDF is actually agnostic to the downstream motion generator. In this work, we instantiate it as a differentiable inter-arm clearance term inside nonlinear model predictive control (NMPC). Compared
with reactive velocity-level collision avoidance based on
simplified geometric proxies~\cite{haviland2021neo}, the
predictive formulation optimizes tracking, regularization,
hardware limits, and inter-arm clearance jointly over a finite
horizon. With the design in Section~\ref{sec:PIUDF Framework}, PI-UDF provides gradient-accessible body-to-body
distance feedback along this horizon, allowing the optimizer
to reason about close-proximity interactions without relying
on inflated proxy geometries.

\subsubsection{NMPC with Differentiable Soft Barriers}
To integrate PI-UDF into a centralized motion planning framework, we define the generic state as $\mathbf{x}_k = [\mathbf{q}_k^\top, \dot{\mathbf{q}}_k^\top]^\top \in \mathbb{R}^{2n}$ and the control input as $\mathbf{u}_k = \ddot{\mathbf{q}}_k \in \mathbb{R}^n$, where \(n=14\) for centralized dual-arm swapping and
\(n=7\) for the single-arm evasion controller.

Evaluating all link pair permutations per integration step is computationally redundant, as many pairs are structurally incapable of colliding. To streamline collision queries, a hierarchical masking strategy can be employed: kinematically impossible pairs are statically pruned, followed by a lightweight broad-phase filter (e.g., bounding capsules) to mask remaining pairs exceeding a safe distance threshold.

However, instantiating a variable number of pairwise constraints would
change the NLP structure across solves. A constant-dimension soft-penalty approach is thus preferable. The filtered pairwise distances are aggregated by a smooth
LogSumExp approximation of the minimum clearance
\(\hat d_{\min}\) over the active pair set \(\mathcal P\):
\begin{equation}
\widetilde{\mathcal D}_{\alpha}(q_k)
=
-\frac{1}{\alpha}
\log
\sum_{(i,j)\in\mathcal P_{\mathrm{act}}}
\exp\!\left(
-\alpha\hat d_{ij}(q_k)
\right),
\label{eq:smooth_min}
\end{equation}
where \(\alpha>0\) controls the approximation sharpness and
\(\mathcal P_{\mathrm{act}}\) denotes the active inter-arm pair
set retained by the broad-phase mask.

Using this smooth minimum, we define an exponential
soft-barrier and solve the following NMPC problem over a
prediction horizon \(N\):
\begin{subequations}
\begin{align}
    \min_{\mathbf{X}, \mathbf{U}} \quad & \sum_{k=0}^{N-1} e^{\gamma k} \Big( \ell_{\text{track}}(\mathbf{x}_k) + \ell_{\text{reg}}(\mathbf{u}_k) \Big) + \ell_{\text{coll}}(\mathbf{x}_k) \\
    \text{s.t.} \quad 
    & \mathbf{x}_{k+1} = f_{\text{dyn}}(\mathbf{x}_k, \mathbf{u}_k) \quad \forall k \in [0, N-1] \label{eq:mpc_dyn}\\
    & \mathbf{x}_{\min} \leq \mathbf{x}_k \leq \mathbf{x}_{\max} \label{eq:mpc_state_bnd}\\
    & \mathbf{u}_{\min} \leq \mathbf{u}_k \leq \mathbf{u}_{\max} \label{eq:mpc_ctrl_bnd}
\end{align}
\label{eq:nmpc_ocp}
\end{subequations}
where
\[
\ell_{\mathrm{coll}}(x_k)
=
w_{\mathrm{coll}}
\exp\!\left[
-\beta\left(
\widetilde{\mathcal D}_{\alpha}(q_k)
-d_{\mathrm{safe}}
\right)
\right].
\]
is a smooth collision penalty around the safety margin
\(d_{\mathrm{safe}}\). The exponential form remains small in
well-separated states and increases rapidly near the boundary,
providing continuous repulsive gradients without the numerical
singularities of logarithmic or inverse barriers. The factor
\(e^{\gamma k}\) emphasizes later-stage tracking and regularization terms, and the state
and control bounds enforce hardware limits.
\subsubsection{Real-Time-Oriented JIT-Compiled NMPC Implementation}
Solving the NMPC problem requires repeated evaluation of
PI-UDF and its derivatives along the prediction horizon.
Directly coupling deep-learning models with external nonlinear
programming solvers can introduce callback latency, host-device
transfer overhead, and Python Global Interpreter Lock (GIL)
bottlenecks, which may limit the achievable planning frequency
in closed-loop control~\cite{salzmann2023neural}.

To reduce these overheads, we port the trained PI-UDF model to
a functional \texttt{JAX} implementation and express the forward
kinematics, PI-UDF inference, soft-min collision aggregation,
and L-BFGS-B optimization loop in a single JAX/XLA pipeline.
The resulting end-to-end JIT-compiled computation graph enables
automatic differentiation through the full NMPC objective and
amortizes runtime overhead across repeated closed-loop solves~\cite{jax2018github,frostig2019compiling}.

In hardware, the compiled planner is paired with a command
interpolation layer that converts each optimized joint trajectory
into high-rate references for the low-level robot controllers.
The task-specific planning periods, workstation specifications,
and controller rates are reported in Sec.~\ref{sec:hardware_setup}.

\section{Real-Robot Experiments}
\subsection{Hardware Setup and Evaluation Protocol}
\label{sec:hardware_setup}
Building on the model evaluation and NMPC formulation
of Secs.~III--IV, we evaluate PI-UDF as a differentiable
safety primitive for closed-loop dual-arm hardware control.
All experiments are conducted on a dual Franka Emika Panda
system with a fixed relative base placement; the two arms
share a common workspace and therefore require continuous
inter-arm collision reasoning during execution. We instantiate
the controller in two forms: a centralized 14-DoF dual-arm
NMPC for tightly coupled swapping, where both robots are
optimized jointly, and a single-arm evasion NMPC, where R1
tracks a task target while avoiding R2. In the evasion tasks,
R2 is treated as an externally generated moving body rather
than a jointly optimized agent. Depending on the operating
mode, future R2 samples are either supplied to the prediction
horizon or R2 is frozen at its current pose.

All hardware controllers run on a centralized workstation
running Ubuntu 22.04 with PREEMPT-RT, equipped with an
AMD Threadripper 3960X CPU and an NVIDIA RTX A5000
GPU. The high-speed dual-arm swapping controller uses a
\(200\,\mathrm{ms}\) planning period and passes the optimized
joint trajectory through a \(100\,\mathrm{Hz}\) command-interpolation
interface, while the low-level Franka controllers run at
\(1\,\mathrm{kHz}\). The dynamic-evasion controllers use a
\(100\,\mathrm{ms}\) planning period with the same
\(100\,\mathrm{Hz}\) interpolation interface. Solve-time variability is treated as an implementation diagnostic rather than a hard real-time guarantee.

In all hardware experiments, PI-UDF is queried online to
provide pairwise inter-arm distances and gradients. The
NMPC objective uses the smooth aggregate
\(\widetilde{\mathcal D}_{\alpha}(q)\) from
Eq.~\eqref{eq:smooth_min}, whereas the plots and tables
report the raw predicted minimum
\(\hat{\mathcal D}(q)\) defined in
Eq.~\eqref{eq:raw_predicted_clearance}.

The nominal operating threshold is set according to the
control architecture. Centralized dual-arm swapping uses
\(d_{\mathrm{safe}}=5\,\mathrm{cm}\), since both arms are jointly
optimized over a self-consistent 14-DoF horizon. Single-arm
dynamic evasion uses the more conservative value
\(d_{\mathrm{safe}}=10\,\mathrm{cm}\), since only \(R_1\) is
optimized while \(R_2\) is externally generated. In both
cases, \(d_{\mathrm{safe}}\) defines the operating clearance of
the differentiable soft cost rather than a hard collision
constraint.

For post-hoc geometry validation, Drake/FCL~\cite{drake}
is evaluated offline on the logged current configurations of
both arms to obtain the reference
\(\mathcal D^\star(q)\); it is never queried by the online
controller. In the dynamic-evasion mode study, future
\(R_2\) samples affect only the NMPC prediction horizon and
are not used in the reported per-time-step
PI-UDF--Drake/FCL clearance comparisons.

The following subsections progress from close-proximity
centralized swapping to sustained dynamic evasion and
representative reactive, predictive, and waypoint-shaped
operating modes.

\subsection{Close-Proximity 14-DoF Dual-Arm Swapping}
\label{sec:hardware_swapping}

We next stress-test PI-UDF in a high-speed close-proximity swapping task. Both Franka arms are controlled by the centralized 14-DoF NMPC and exchange sides through the same interaction region under the tight threshold \(d_{\mathrm{safe}}=5\,\mathrm{cm}\), without relying on Drake/FCL online.

\begin{figure}[tbp]
  \centering
  \includegraphics[width=0.98\linewidth]{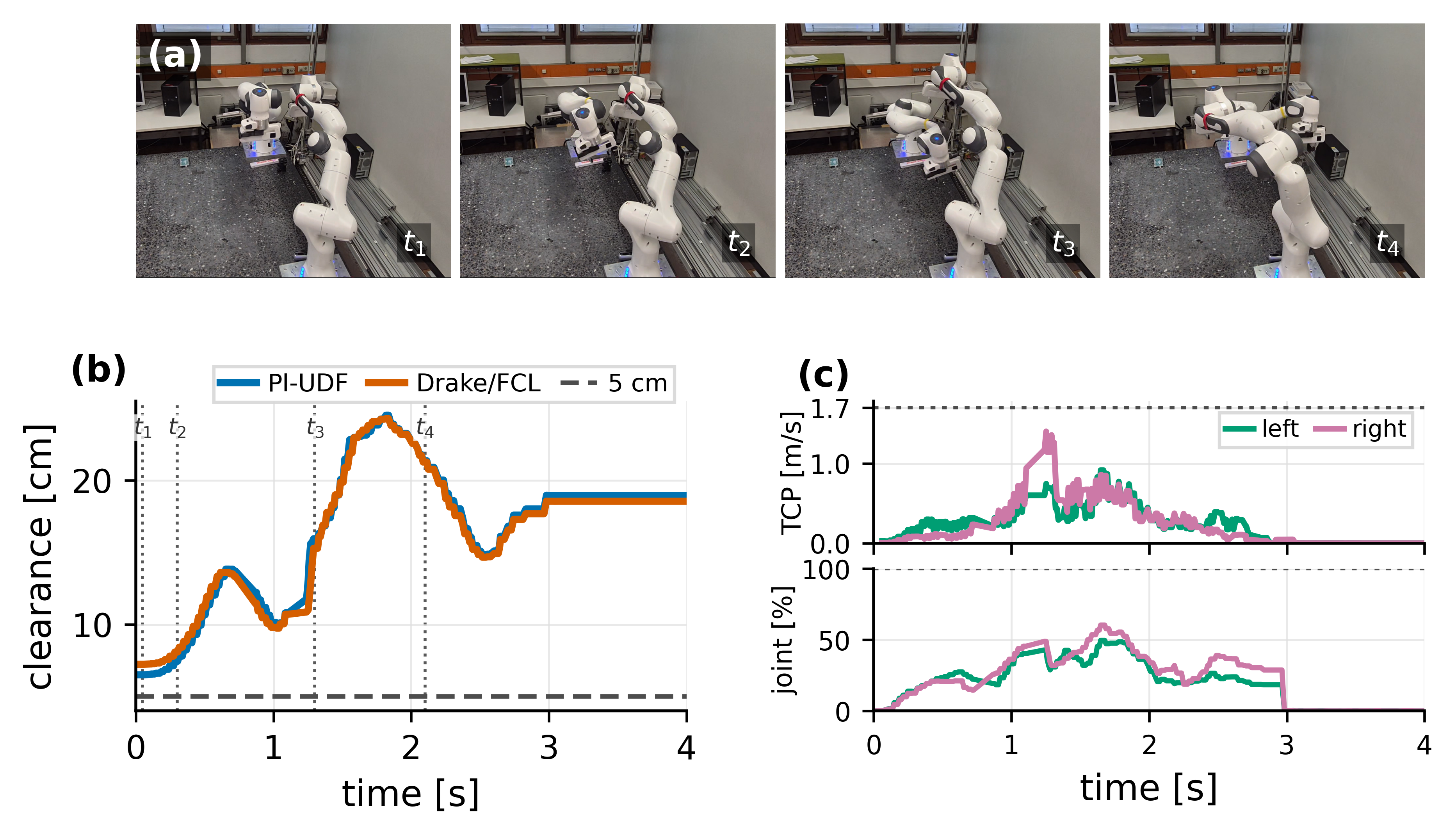}
  \caption{
  High-speed close-proximity dual-arm swapping.
  (a) Representative frames from the executed sequence showing the system configurations.
  (b) Raw PI-UDF minimum
  \(\hat{\mathcal D}(q_{\mathrm{real}})\) and offline Drake/FCL
  clearance \(\mathcal D^\star(q_{\mathrm{real}})\) evaluated on
  the executed two-arm trajectory.
  (c) TCP speeds and per-arm max joint-speed ratios.
  }
\label{fig:highspeed_swapping}
\end{figure}

Fig.~\ref{fig:highspeed_swapping} summarizes the executed run. Fig.~\ref{fig:highspeed_swapping}(a) shows four representative frames \(t_1\)--\(t_4\) from the active swapping motion, illustrating the two arms entering, crossing, and leaving the shared interaction region. Fig.~\ref{fig:highspeed_swapping}(b) reports the active \(0\)--\(4\,\mathrm{s}\) clearance window. The raw PI-UDF clearance
\(\hat{\mathcal D}(q_{\mathrm{real}})\) and the Drake/FCL
reference \(\mathcal D^\star(q_{\mathrm{real}})\) are evaluated
on the same executed two-arm trajectory. This executed-state replay provides an apples-to-apples geometry comparison. The minimum logged PI-UDF clearance is
\(6.04\,\mathrm{cm}\), while offline Drake/FCL replay gives
\(7.24\,\mathrm{cm}\); neither trace falls below the
\(5\,\mathrm{cm}\) threshold. The active motion settles after approximately \(3.0\,\mathrm{s}\), while the closest interaction occurs during the first second of the run.

Fig.~\ref{fig:highspeed_swapping}(c) provides the related motion-speed context. The TCP-speed traces show a rapid transient during the crossing phase. The upper plot reports, for each arm, the maximum normalized joint speed over its seven joints,
\[
r_{\max}(t)=\max_j \frac{|\dot q_j(t)|}{\dot q_{j,\max}},
\]
which peaks at approximately \(50\%\) and \(60\%\) for the two arms. Thus, the swapping motion is not a quasi-static avoidance maneuver, while still remaining below joint-speed saturation. These speed traces contextualize the difficulty of the executed motion.

\subsection{Sustained Single-Arm Dynamic Evasion}
\label{sec:dynamic_evasion_baseline}
We next move from centralized dual-arm NMPC to a single-arm dynamic-evasion setting, where R1 is optimized by NMPC while R2 is treated as an external moving robot body. In this baseline run, R2 follows a scripted trajectory composed of multiple motion segments, and R1 replans to track a task target while maintaining inter-arm clearance. This setting introduces the dynamic-evasion architecture used in the following operating-mode study.

\begin{figure}[tbp]
  \centering
  \includegraphics[width=0.98\linewidth]{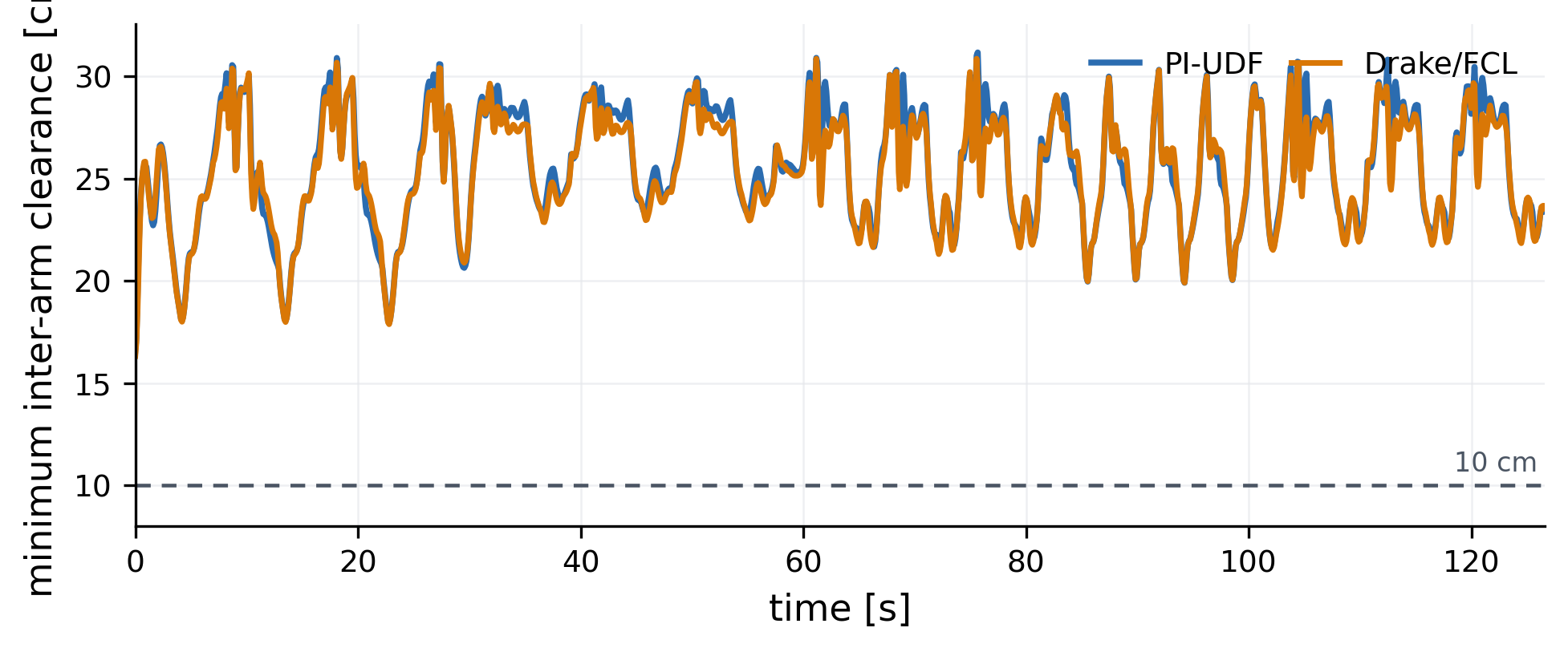}
  \caption{Raw logged PI-UDF clearance \(\hat{\mathcal D}(q_t)\) and
  offline Drake/FCL replay \(\mathcal D^\star(q_t)\) along the
  logged sustained dynamic-evasion trajectory.}
\label{fig:dynamic_evasion_baseline_clearance_full}
\end{figure}

Fig.~\ref{fig:dynamic_evasion_baseline_clearance_full} shows a 126.6 s run under a 10 cm nominal clearance threshold. The raw predicted clearance
\(\hat{\mathcal D}\) remains above the threshold throughout the run, with a minimum clearance of 16.56 cm. Offline Drake/FCL replay gives
\(\min_t\mathcal D^\star(q_t)=16.33\,\mathrm{cm}\), also with zero samples below 10 cm. This run serves as a sustained-operation check for the single-arm dynamic-evasion architecture; the next subsection studies representative operating modes that vary how R2 motion is incorporated into the MPC horizon.

\subsection{Dynamic-Evasion Operating Modes}
\label{sec:dynamic_evasion_modes}
\begin{figure}[tbp]
  \centering
  \includegraphics[width=0.98\linewidth]{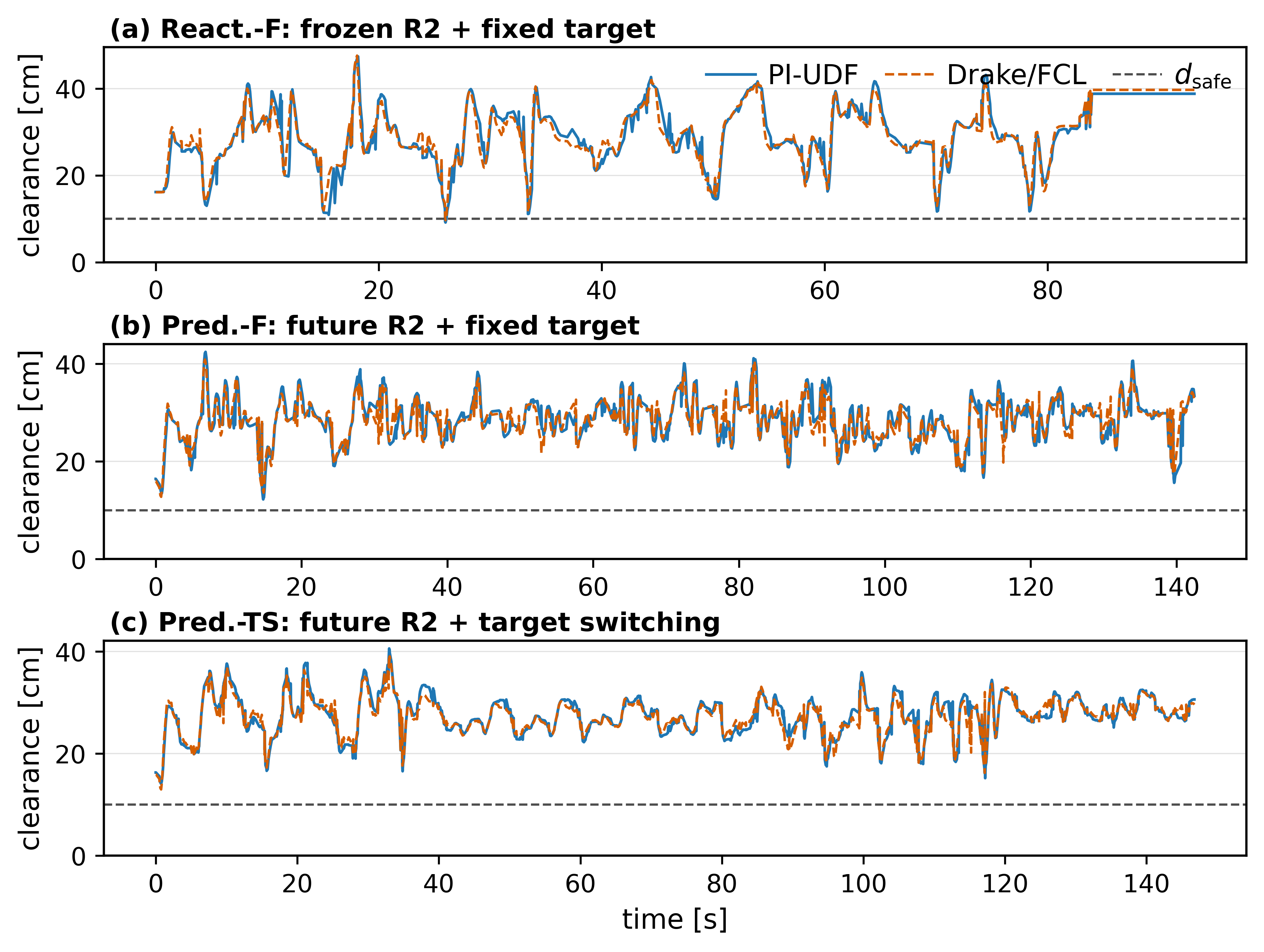}
  \caption{
  Raw logged PI-UDF clearance \(\hat{\mathcal D}(q_t)\) is
  compared with offline Drake/FCL clearance
  \(\mathcal D^\star(q_t)\) evaluated at the same current
  logged configurations.
  }
\label{fig:dynamic_evasion_modes}
\end{figure}

We finally evaluate the PI-UDF-NMPC framework under several real-time motion-planning configurations for dynamic evasion. The goal is not to isolate a single MPC design choice, but to test whether the learned differentiable distance field remains reliable when integrated with different horizon and task-target formulations. In particular, the moving R2 arm can be treated as a frozen obstacle in the horizon, predicted over the horizon using its scripted future trajectory, or combined with online target switching for R1 as the interaction region changes. These differences affect how the
NMPC predicts and optimizes future motion; the clearance
values reported in Table~\ref{tab:dynamic_evasion_modes}
and Fig.~\ref{fig:dynamic_evasion_modes} are evaluated at
the current logged system configurations.

\begin{table}[tbp]
\centering
\caption{
Dynamic-evasion planning configurations and clearance outcomes.
}
\label{tab:dynamic_evasion_modes}
\footnotesize
\setlength{\tabcolsep}{3pt}
\begin{tabular}{@{}l l c c@{}}
\toprule
Mode & Planning configuration & Min. clr. [cm] & Below \(d_{\mathrm{safe}}\) \\
& & \(\hat{\mathcal D}/\mathcal D^\star\) & \(\hat{\mathcal D}/\mathcal D^\star\) \\
\midrule
React.-F
& frozen R2 + fixed target
& 9.16 / 9.68
& 3 / 1 \\
Pred.-F
& future R2 + fixed target
& 12.22 / 13.93
& 0 / 0 \\
Pred.-TS
& future R2 + target switching
& 14.09 / 15.91
& 0 / 0 \\
\bottomrule
\end{tabular}
\end{table}

Table~\ref{tab:dynamic_evasion_modes} summarizes these
configurations and their current-state clearance outcomes.
React.-F freezes the current R2 configuration across the
MPC horizon while R1 tracks a fixed target. Pred.-F
supplies the scripted future R2 trajectory to the horizon,
also with a fixed R1 target. Pred.-TS uses the same
predictive R2 horizon together with task-level target
switching for R1 as the interaction region changes.
Since these planning configurations differ not only in horizon
information but also in task-target policy and execution
details, they should be interpreted as representative
PI-UDF-NMPC integration tests rather than as a controlled
ablation of prediction alone.

Fig.~\ref{fig:dynamic_evasion_modes} compares the logged PI-UDF clearance with offline Drake/FCL replay evaluated on the same current logged two-arm configurations. This same-\(q\) comparison checks whether the learned distance field remains geometrically consistent across the three planning configurations. Although Pred.-F and Pred.-TS use
future \(R_2\) samples inside the NMPC horizon, those
future samples are not used to compute the per-time-step
clearance traces shown here.

The React.-F run provides the closest interaction in this set of tests. It falls below the 10 cm nominal threshold for only a few MPC-rate samples: three samples according to the logged PI-UDF clearance and one sample according to recorded-state FCL replay. Its minimum FCL clearance is 9.68 cm, indicating a short violation of the nominal soft margin rather than a physical collision. Both predictive configurations remain above the threshold in the recorded-state replay. Across the tested operating modes, PI-UDF provides consistent current-state clearance estimates while supporting frozen-R2, predictive-R2, and target-switching NMPC formulations.

\section{Conclusion}
This work presented PI-UDF, a compact differentiable
body-to-body clearance model that combines FK-derived
relative poses, task-optimized link embeddings, and
safety-aware boundary learning. Model-level experiments
showed competitive global accuracy, the lowest FNR among
the compared models, and substantially lower training cost
than the re-trained PairwiseNet baseline.

Integrated into NMPC, PI-UDF supported high-speed
close-proximity 14-DoF swapping and sustained dynamic
evasion on real dual-Franka hardware. Offline Drake/FCL
replay confirmed the reported current-state clearance
behavior. Future work will address uncertainty-aware motion
prediction, solver robustness, and generalization across
robot layouts and morphologies. The use of PI-UDF as a geometry prior or safety critic for learning-based policy and trajectory generators can be interesting as well.

\bibliographystyle{ieeetr}
\bibliography{references}

\end{document}